\pdfoutput=1
\documentclass[11pt]{article}

\usepackage{ACL2023}

\usepackage{times}
\usepackage{latexsym}
\usepackage{graphicx}       % graphics
\usepackage{subcaption} % for subfigures

\usepackage[T1]{fontenc}
\usepackage[utf8]{inputenc}

\usepackage{microtype}

\usepackage{inconsolata}
\usepackage{listliketab,tabularx,booktabs}

\title{Environmental Claim Detection}

\author{Dominik Stammbach \\
  ETH Zurich \\
  \texttt{dominsta@ethz.ch} \\\And
  Nicolas Webersinke \\
  FAU Erlangen-Nuremberg \\
  \texttt{nicolas.webersinke@fau.de} \\\And
  Julia Anna Bingler \\ 
  Council on Economic Policies \\
  \texttt{jb@cepweb.org} \\\AND
  Mathias Kraus \\
  FAU Erlangen-Nuremberg \\
  \texttt{mathias.kraus@fau.de} \\\And
  Markus Leippold \\
  University of Zurich \\
  \texttt{markus.leippold@bf.uzh.ch} \\
}

\begin{document}
\maketitle
\begin{abstract}
To transition to a green economy, environmental claims made by companies must be reliable, comparable, and verifiable. To analyze such claims at scale, automated methods are needed to detect them in the first place. However, there exist no datasets or models for this. Thus, this paper introduces the task of environmental claim detection. To accompany the task, we release an expert-annotated dataset and models trained on this dataset. We preview one potential application of such models: We detect environmental claims made in quarterly earning calls and find that the number of environmental claims has steadily increased since the Paris Agreement in 2015.
\end{abstract}

\section{Introduction}

In the face of climate change, we witness a transition towards a more sustainable and green economy. This change is driven by changes in regulation, public opinion, and investor attitudes. For example, global assets managed under a sustainability label are on track to exceed \$53 trillion by 2025, more than a third of total assets under management. However, unfortunately, the boom has been accompanied by rampant greenwashing, with companies boasting about their environmental credentials.\footnote{See, e.g., The Economist, May 22nd, 2021.} % made greenwashing argument
Because of this surge in environmental claims and to protect consumers, initiatives on substantiating green claims are developed.\footnote{For example an EU initiative on green claims:  \url{https://ec.europa.eu/environment/eussd/smgp/initiative_on_green_claims.htm}} Due to an ever-growing amount of text, there is a need for automated methods to detect environmental claims. Detecting such claims at scale can assist policy-makers, regulators, journalists, activists, the research community, and an informed public in analyzing and scrutinizing environmental claims made by companies and facilitating the transition to a green economy.

\begin{figure}
    \centering
    \footnotesize
    \begin{tabular}{p{7.5cm}}
    \textbf{Environmental claim}: A total population of 6148 is getting the benefit of safe potable drinking water due to this initiative. \\
    \textbf{Environmental claim}: Hydro has also started working on several initiatives to reduce direct CO2 emission in primary aluminium production. \\
    \textbf{Negative example}: Generally, first of all, our Transmission department is very busy, both gas and electric transmission, I should say, meeting the needs of our on-network customers. \\
    \textbf{Negative example}: Teams are thus focused on a shared objective in terms of growth and value creation. \\
    \end{tabular}
    \caption{Environmental Claims and Negative Examples from our dataset.}
    \label{fig:examples}
\end{figure}

Thus, we introduce the task of environmental claim detection. Environmental claim detection is a sentence-level classification task with the goal of predicting whether a sentence contains an environmental claim or not. Often, environmental claims are made in a clear and concise matter on a sentence level, with the intention to convey to a consumer or stakeholder that a company or product is environmentally friendly. 

To facilitate future research on environmental claim detection, we release an expert-annotated dataset containing real-world environmental claims and models which can be used by practitioners. For constructing the dataset, we were inspired by the European Commission (EC), which defines such claims as follows: \textit{Environmental claims refer to the practice of suggesting or otherwise creating the impression (in the context of a commercial communication, marketing or advertising) that a product or a service is environmentally friendly (i.e., it has a positive impact on the environment) or is less damaging to the environment than competing goods or services}.\footnote{From the Commission Staff Working Document, Guidance on the implementation/application of Directive 2005/29/EC
on Unfair Commercial practices, Brussels, 3 December 2009 SEC(2009) 1666. See section 2.5 on misleading
environmental claims.} While such claims can be truthful and made in good faith, boasting about environmental credentials can also be monetized \cite{greenwashing_monetized}. For example, consumers are willing to spend more money on environmentally friendly products \cite{nielsen_media_research}. The Commission states if environmental claims are too vague, unclear, or misleading, we are confronted with an instance of "greenwashing" (this definition is given in the same Commission Staff Working Document).

We situate environmental claim detection at the intersection of claim detection \citep[e.g.,][]{arslan_claimbuster} and pledge detection \cite{pledge_specificity, pledge_specificity_2}. An environmental claim is typically made to increase the environmental reputation of a firm or a product. We show that models trained on the current claim and pledge detection datasets perform poorly at detecting environmental claims, hence the need for this new dataset. We make our dataset, code and models publicly available.\footnote{We host all code, data and models on \url{https://github.com/dominiksinsaarland/environmental_claims}. The dataset can also be accessed as a \href{https://huggingface.co/datasets/climatebert/environmental_claims}{hugginface dataset}, and our model is available on the \href{https://huggingface.co/climatebert/environmental-claims}{huggingface model hub}.} Lastly, we envision computer-assisted detection of greenwashing in future work, i.e., the automatic determination if an environmental claim is false, too vague, non-verifiable, or misleading. To make progress on automated greenwashing detection, it is mandatory to first detect environmental claims at scale.

\section{Related Work}

This work is part of an ongoing effort at the intersection of environmental and climate change-related topics and natural language processing \cite{stede-patz-2021-climate}. Resulting datasets and methods can help regulators, policy-makers, journalists, the research community, activists, and an informed public investigate such topics at scale with the help of computer assistance. Methods include ClimateBERT \cite{climatebert}, and ClimateGPT \cite{climateGPT}, two language models pre-trained on climate-related text. NLP tasks and datasets include climate change topic detection \cite{climatetext} and detecting media stance on global warming \cite{luo-etal-2020-detecting}. \citet{one_tweet_at_a_time} collect climate change opinions at scale from social platforms, \citet{Al-Rawi2021-to} analyze fake news Tweets around climate change. In a similar direction \citet{Coan2021} analyze contrarian claims about climate change and \cite{piskorski-2022-exploring} explore data augmentation techniques for climate change denial classification. Further, there exists work about claim verification of climate change related claims \cite{leippold2020climatefever}, detecting media stance on global warming \cite{luo-etal-2020-detecting}, collecting climate change opinions at scale from social platforms \cite{one_tweet_at_a_time}, and finally, the analysis of regulatory disclosures \cite{friedrich_et_al, kolbel2020ask}.

In this broader context of applying NLP methods for climate change-related topics, We situate environmental claim detection at the intersection of claim spotting and pledge detection, covering the domain of text produced by companies with the goal of boosting their environmental credentials. Claim spotting is the task of finding fact-check worthy claims \cite{arslan_claimbuster, clef_2018, clef_2020}. Pledge detection aims to detect pledges made in, for example, political campaigns \cite{pledge_specificity, pledge_specificity_2}. Environmental claims state an environmental benefit (claim) or convey the intention (pledge) for a material impact, i.e., some environmental benefit, which pleases the audience  (consumers or stakeholders) of the claim.

\section{Dataset}

Our dataset contains environmental claims made by listed companies. We collected text from sustainability reports, earning calls, and annual reports of listed companies and annotated 3'000 sentences. After discarding tied annotations, our resulting dataset contains 2'647 examples.\footnote{In the GitHub repository, we also include a link to all 3'000 sentences, with the 4 individual annotations for each datapoint (including ties), in case this additional information is useful for follow-up research.} We provide dataset statistics in Table \ref{tab:dataset_stats} and a text length histogram in Appendix Table \ref{tab:app_histogram_textlength}.

\begin{table}[t!]
    \centering
    \begin{tabular}{c c c c }
    split & \# examples & mean length & claims (\%) \\ \midrule
    train & 2117 & 24.4 & 0.25 \\ 
    dev & 265 & 24.2 & 0.25 \\ \
    test & 265 & 24.9 & 0.25 \\ 
    all & 2647 & 24.5 & 0.25 \\ \midrule
        \end{tabular}
    \caption{Dataset Statistics}
    \label{tab:dataset_stats}
\end{table}

\begin{table*}[t!]
    \centering \small{
    \begin{tabular}{l | c c c c | c c c c | c c c c}
        model & pr & rc & F1 & acc & pr & rc & F1 & acc & pr & rc & F1 & acc \\
        &     \multicolumn{4}{c}{CV} &     \multicolumn{4}{c}{dev} &     \multicolumn{4}{c}{test} \\ \hline
        Majority baseline & 0.0 & 0.0 & 0.0 & 74.9 & 0.0 & 0.0 & 0.0 & 74.7 & 0.0 & 0.0 & 0.0 & 75.1 \\
        Random baseline & 26.2 & 53.2 & 35.1 & 50.5 & 27.9 & 58.2 & 37.7 & 51.3 & 26.2 & 46.6 & 33.5 & 53.5 \\
        ClaimBuster RoBERTa  & 27.9 & 62.6 & 38.6 & 49.9 & 27.3 & 52.7 & 35.9 & 47.5 & 25.3 & 51.4 & 33.9 & 45.7 \\
        Pledge Detection RoBERTa & 26.2 & 31.7 & 28.7 & 60.4 & 27.6 & 28.4 & 28.0 & 59.2 & 24.1 & 29.2 & 26.4 & 55.8 \\ \midrule
        TF-IDF SVM & 71.1 & 65.9 & 68.4 & 84.7 & 67.7 & 63.6 & 65.6 & 83.4 & 68.1 & 70.1 & 69.1 & 84.2 \\
        Character n-gram SVM & 76.8 & 63.6 & 69.6 & 86.0 & 69.2 & 68.2 & 68.7 & 84.5 & 75.0 & 67.2 & 70.9 & 86.0 \\
        DistilBERT & 79.9 & 89.0 & 84.2 & 91.6 & \underline{77.5} & \textbf{93.9} & \underline{84.9} & \underline{91.7} & 74.4 & \textbf{95.5} & 83.7 & 90.6 \\  
        ClimateBERT & \underline{80.1} & \underline{90.1} & \underline{84.8} & \underline{91.9} & 76.9 & 90.9 & 83.3 & 90.9 & \underline{76.5} & 92.5 & \underline{83.8} & \underline{90.9} \\       
        RoBERTa\textsubscript{base} &  77.8 & \textbf{91.3} & 84.0 & 91.3 & 74.7 & \textbf{93.9} & 83.2 & 90.6 & 73.3 & \underline{94.0} & 82.4 & 89.8 \\
        RoBERTa\textsubscript{large} &  \textbf{83.1} & \underline{90.1} & \textbf{86.4} & \textbf{92.9} & \textbf{80.5} & \textbf{93.9} & \textbf{86.7} & \textbf{92.8} & \textbf{78.5} & 92.5 & \textbf{84.9} & \textbf{91.7} \\ \midrule       \end{tabular} }
    \caption{Main results: We report precision, recall, F1, and accuracy on a cross-validation split (CV), the development set (dev), and the test set of the environmental claims dataset. All numbers are reported as \%, and best performance per split is indicated in bold, the second best is underlined.}
    \label{tab:experiments}
\end{table*}

The dataset is annotated by 16 domain experts.\footnote{All annotators passed a core course on sustainable investing with a high grade. This course is part of the executive education program for the Master of Advanced Studies in Sustainable Finance, offered by the University of Zurich. Most of the annotators have prior work experience in the financial sector.} The authors drafted annotation guidelines in an iterative process and added examples of clear and borderline environmental claims to the guidelines. In Appendix \ref{sec:annotation_guidelines}, we list the complete guidelines available to the annotators, along with examples and rationales that the authors discussed in pilot annotation rounds. %Employing domain experts, a moderate inner annotator agreement, and carefully drafted annotation guidelines lead us to believe that our dataset is of high quality.  

To extract the sentences annotated in our dataset, we use a preliminary model to sample candidate sentences from various text sources produced by firms. Furthermore, we randomly sample sentences from different clusters obtained with k-means to increase the coverage of the domain. We describe the sampling process of the dataset in detail in Appendix \ref{sec:sample_selection} and provide further information on the data sources in Appendix \ref{app:data_sources}.

While we do not release a large-scale dataset, this is the result of a conscious decision to prioritize quality over quantity. We employed domain experts to annotate the data, which results in costly annotations. In Appendix  \ref{sec:dataset_size}, we show that the performance of models converges after being trained on more than 60\% of the training set, and we find diminishing marginal utility of including more sentences. Hence our decision to stop annotation here and release an annotated dataset with 2'647 examples.

We assigned each sentence to four annotators. The annotations are aggregated by majority vote. 60\% of the 3'000 samples was decided unanimously by the annotators, and 88.3\% of the annotations made were part of a majority decision. 353 sentences received tied annotations (11.7\% of the samples), and we discarded these examples from the dataset.%\footnote{We also host all 3'000 examples, including ties, in our \href{https://github.com/dominiksinsaarland/environmental_claims}{github repository}.}
The overall inter-annotator agreement measured in Krippendorff's alpha is 0.47, indicating moderate agreement. 

\section{Experiments}\label{sec:experiments}

We conduct two types of experiments: (1) We examine the performance of various models on our dataset, among them pre-trained claim and pledge detection models and fine-tuned environmental claim detection transformer models \citep[such as, e.g.][]{devlin-etal-2019-bert, roberta, Sanh2019DistilBERTAD, climatebert}. (2) we apply our models to the text produced by listed companies, which leads to a small case study demonstrating one of the intended use cases of the dataset. 

\subsection{Environmental Claim Detection Models}

\begin{figure*}[t!]
    \centering
    \includegraphics[width=0.7\linewidth]{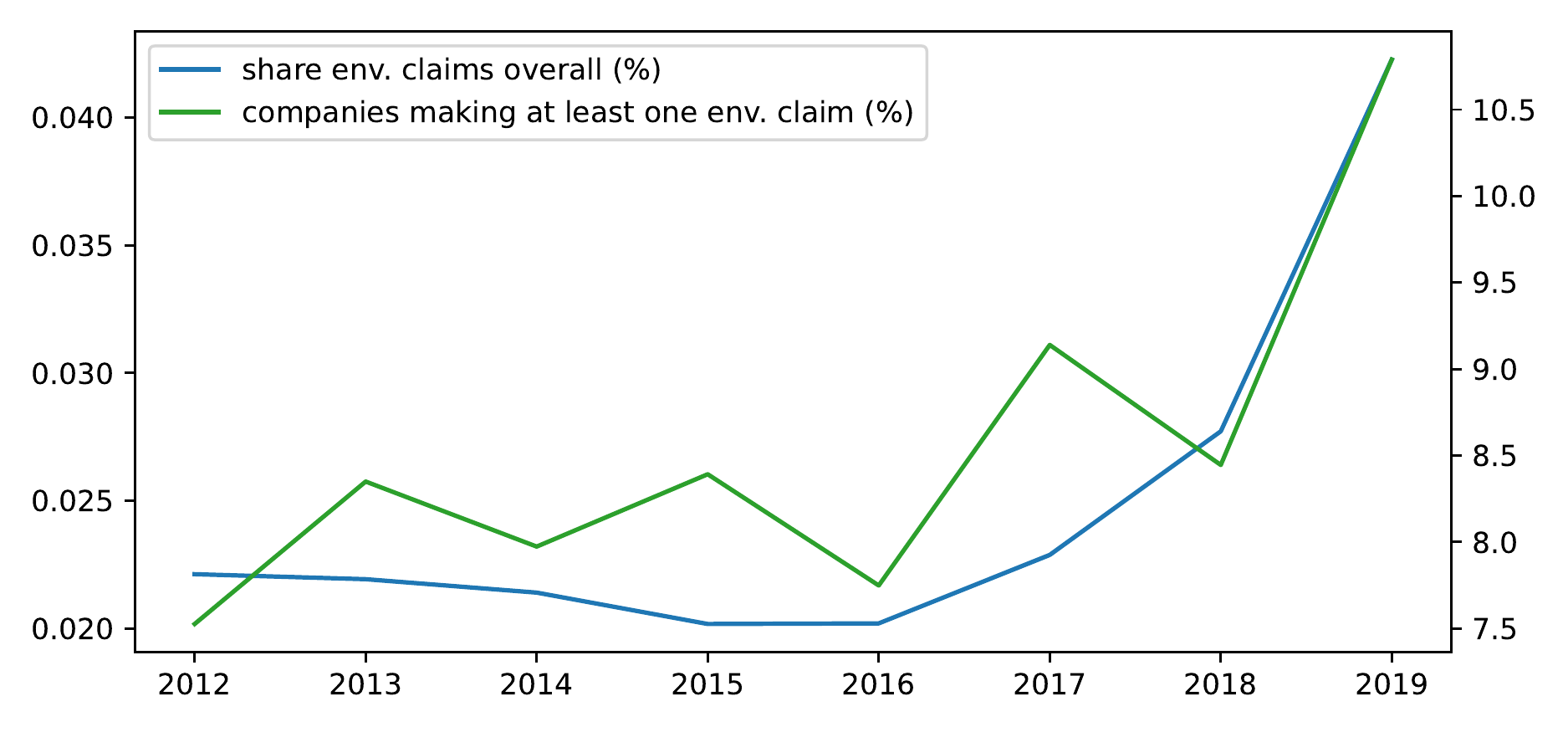}
    \caption{Amount of environmental claims (in \%) made in earning calls answer sections. The blue line (y-axis on the left) shows the share of environmental claims made each year. The green line shows the share of companies making at least one environmental claim in a given year (y-axis on the right).}
    \label{fig:earning_calls}
\end{figure*}

We report various metrics on a 5-fold cross-validation split of the whole dataset, the development, and the test set in Table \ref{tab:experiments}. We present two poorly performing baselines: \textit{majority}, where we assign the not-a-claim label to all examples, and \textit{random}, where we randomly assign one of the two labels to each example. Next, we fine-tune a RoBERTa\textsubscript{base} model on the ClaimBuster dataset \cite{arslan_claimbuster}, and use this model to detect environmental claims in the dataset.\footnote{We train the model to distinguish fact-check-worthy claims vs. all other claims. The model works exceptionally well on the ClaimBuster test set with a micro-F1 of 97.9\% and a macro-F1 of 97.0\%.} While achieving rather high recall, the model does not cope well with the domain shift and fails to detect environmental claims reliably. Similar findings hold for a RoBERTa\textsubscript{base} model trained on a Pledge Detection dataset \cite{pledge_specificity}.\footnote{The model achieves a 67\% F1 score and 78\% accuracy on a held-out split of the Pledge Detection but also fails to adapt to detect environmental claims.} These results highlight the need for a dedicated dataset.

Furthermore, we train two SVM models. The first one uses tf-idf bag-of-word features, the second is based on character n-gram features. Both models achieve an acceptable F1 score between 65\% and 71\% on all dataset splits, indicating that using environment-related keywords or n-grams is indicative of whether a sentence is an environmental claim. However, all transformer models explored in this study outperform the SVM, hence the presence of environmental keywords alone is not sufficient for predicting such claims. Especially for recall, we find a large gap between transformer and SVM models of up to 25\% points. We interpret this gap as evidence that not all environmental claims contain distinguishing environmental keywords.

Lastly, we fine-tune various transformer models \cite{roberta, Sanh2019DistilBERTAD, climatebert}. They all achieve an F1 score higher than 82\% on all different dataset splits, a vast performance increase compared to the other models examined so far. We observe only minor differences between these models. The biggest model RoBERTa\textsubscript{large} achieves the best scores overall, followed by ClimateBERT, a DistilBert-like language model further pre-trained on over 1.6 million climate-related paragraphs. Hence, further pre-training on climate-related text seems beneficial to detect environmental claims.

For training our models, we use Hugging Face \cite{wolf-etal-2020-transformers} and standard RoBERTa hyper-parameters. We use the Adam optimizer with a learning rate of 2e-5, a batch size of 16, and train models for 3 epochs. To minimize compute and environmental footprint of our experiments and due to consistent results over different dataset splits, we did not explore other hyper-parameters in more detail and reported only results of single runs.

\subsection{Earning Calls}

\begin{figure*}
    \centering
    \includegraphics[width=0.475\linewidth]{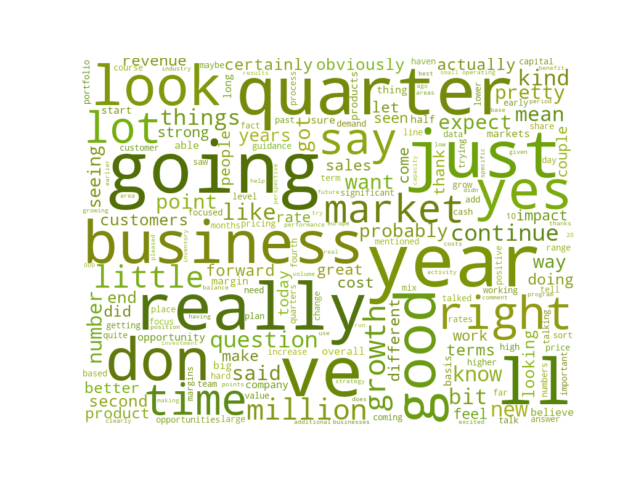}
    \includegraphics[width=0.475\linewidth]{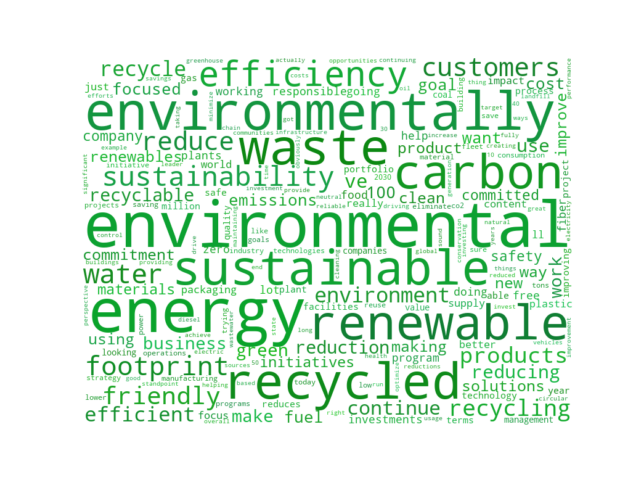} 
    \caption{Word clouds of non-claims (on the left) and environmental claims (on the right) in earnings call transcripts.}
    \label{fig:wordclouds}
\end{figure*}

We use our trained model to detect environmental claims in corporate earning calls between 2012 and 2020. These are conference calls between the management of a publicly traded company, analysts, investors, and the media to discuss the company's financial results and other topics for a given reporting period (mainly quarterly). The conference calls consist of different segments, of which the segment with questions and answers is the most interesting for our purposes. Therefore, we focus on the management responses, which consist of 12 million sentences from 3,361 unique companies. All earnings conference call transcripts are obtained from Refinitiv Company Events Coverage. Due to the size of the data and computational constraints, we use our ClimateBERT model, fine-tuned on detecting environmental claims instead of the RoBERTa\textsubscript{large} model. 

We would expect that the amount of environmental claims made by corporations and business leaders has steadily increased since the Paris Agreement in 2015. In Figure \ref{fig:earning_calls}, we find that this is indeed the case. The amount of environmental claims is not only increasing, but the increase is also accelerating. In 2019, the share of environmental claims is twice as high as in 2015. Not only the amount of environmental claims made in earning calls is increasing, but also the share of companies who makes such claims increased by 33\%, and in 2019, one in ten companies makes at least one environmental claim in the answer sections of an earning call.

In Figure \ref{fig:wordclouds}, we display word clouds for the most important words classified as non-claims (on the left), and the most important words for environmental claims (on the right). It is evident that the sentences classified as claims contain more environmental-related keywords; We see that these keywords cover different environmental aspects, e.g., recycling and waste, carbon and emissions, renewables, water, etc. In Appendix Table \ref{fig:examples_earning_calls}, we additionally list the 5 highest and lowest scoring sentences based on our model. Our model effectively identifies environmental claims as the predominant category at the upper end of the distribution, whereas it appears that such claims are absent in the lower end of the distribution.

This small case study illustrates one of the intended use cases of our dataset and the associated models: We present a tool that allows us to detect environmental claims at scale. Having access to environmental claims at scale makes it possible to analyze and scrutinize them in future work.

\section{Conclusion}

The vast and ever-growing volume of corporate disclosures, regulatory filings, and statements in the news calls for an algorithmic approach to detect environmental claims made by companies at scale. Thus, we introduce the NLP task of detecting environmental claims, a dataset containing such claims and associated models which can detect these claims in the wild. Our dataset is annotated by domain experts and thus of high quality. We describe the dataset and its construction process and present various models for detecting environmental claims in our dataset and a small case study.

We envision several directions for future work. First, we plan to investigate "greenwashing", the practice of making a false, vague, unclear, or misleading environmental claim. To make progress on this front, it is mandatory that we can detect environmental claims in the first place. Second, models trained on detecting environmental claims have merits of their own, as previewed in our case study. We plan to explore more such applications in detail, e.g., analyzing annual reports and TCFD\footnote{Task Force on Climate-Related Financial Disclosures} reports at scale. For example, it would be interesting to see in which sections of TCFD reports firms to make environmental claims. 
Lastly, we expect an increase of contributions at the intersection of environmental topics, climate change, and NLP in the near future. This work contributes to such efforts.

\section*{Limitations}
% EACL 2023 requires all submissions to have a section titled ``Limitations'', for discussing the limitations of the paper as a complement to the discussion of strengths in the main text. This section should occur after the conclusion, but before the references. It will not count towards the page limit.

% The discussion of limitations is mandatory. Papers without a limitation section will be desk-rejected without review.
% ARR-reviewed papers that did not include ``Limitations'' section in their prior submission, should submit a PDF with such a section together with their EACL 2023 submission.

% While we are open to different types of limitations, just mentioning that a set of results have been shown for English only probably does not reflect what we expect. 
% Mentioning that the method works mostly for languages with limited morphology, like English, is a much better alternative.
% In addition, limitations such as low scalability to long text, the requirement of large GPU resources, or other things that inspire crucial further investigation are welcome.

We find several limitations in this work. First, we acknowledge that the technical novelty of this work is limited: We introduce a sequence classification task, and we investigate rather standard models in our experiment section (i.e.,  state-of-the-art transformer language models). Nevertheless, we believe that there is a gap in the literature for the task presented in this work, hence our introduction of the environmental claim detection task, the dataset, and models.

Second, we collect data from sustainability reports, earning calls, and annual reports. However, this does not cover the universe of text where environmental claims are made, e.g., company websites and product descriptions. Also, environmental claims can be made about environmental improvements on a wide range of topics such as carbon emissions, water pollution, and recycling, among others. We discussed creating different datasets, where each dataset is dedicated to one specific issue. However, we leave this to future work. Third, sometimes it is necessary to have access to more context to determine whether a sentence is an environmental claim. We discussed whether it would be beneficial to annotate whole paragraphs instead. However, the trade-off would be exploding annotation work and costs, hence our decision to introduce environmental claims as a sentence-level classification task (and we specifically asked annotators to reject ambiguous cases as environmental claims). Nevertheless, given a unlimited budget, we would have pursued annotating whole paragraphs instead (or annotating all environmental claims in a paragraph).

Our data sources, e.g., sustainability reports, are mostly published by European and US-listed companies, which is reflected in our dataset. We crawled these reports from the SEC\footnote{\url{https://www.sec.gov/}}, hence our dataset contains mostly claims made by (a) big firms and (b) firms from developed countries. It is conceivable that smaller firms and firms from non-developed countries make different environmental claims, and models trained on our dataset might not be suitable to detect these claims.

Moreover, our work is subject to all concerns raised in the Ethics Statement below. We find it important to keep all these perspectives in mind when reading and discussing our work.

\section*{Ethics Statement}

\paragraph{Intended Use:} This dataset will benefit journalists, activists, the research community, and an informed public analyzing environmental claims made by listed companies at scale. Also, we see this as a first step towards algorithmic greenwashing detection using NLP methods. It might also be useful to policy-makers and regulators in both the financial sector and the legal domain. Next, we hope companies are inspired by our work to produce more carefully drafted environmental claims. To conclude, we envision that the dataset and related models bring a large positive impact by encouraging truly environmentally friendly actions and less verbose boasting about environmental credentials. 

\paragraph{Misuse Potential:} Although we believe the intended use of this research is largely positive, there exists the potential for misuse. For example, it is possible that for-profit corporations will exploit AI models trained on this dataset while drafting environmental claims. 

\paragraph{Model Bias:} Although the performance of NLP models usually achieves an F1 score of above 80\%, it is widely known that ML models suffer from picking up spurious correlations from data. Furthermore, it has been shown that large pre-trained language models such as ClimateBERT suffer from inherent biases present in the pre-training data leading to biased models -- and we believe our models presented in this work also suffer from these biases.

%\paragraph{Annotator Bias:} We acknowledge that our annotation guidelines are subject to a specific cultural context and to EU guidelines, which themselves are subject to a specific cultural context. Our annotators are also subject to such a context. Furthermore, our data considered is mostly coming form big companies. We do not rule out that all of these are artifacts in the dataset creation process, and reproducing our study in completely different settings might lead to

\paragraph{Data Privacy:} The data used in this study are mostly public textual data provided by companies and public databases. There is no user-related data or private data involved.

\paragraph{Annotator Salary:} We paid standard research assistant salaries of around \$30 per hour, which is common practice at the University of Zurich. We were upfront in disclosing  to annotators that their annotations will lead to a dataset and models which can automatically detect environmental claims. We found that this goal motivated annotators. We speculate (and hope) annotators interpreted the dataset creation process and the goal of releasing the resulting dataset and models as an AI4Good application. The feedback was overwhelmingly positive, and many annotators have asked whether it is possible to participate in follow-up annotation work related to greenwashing detection. 

%Scientific work published at EACL 2023 must comply with the \href{https://www.aclweb.org/portal/content/acl-code-ethics}{ACL Ethics Policy}. We encourage all authors to include an explicit ethics statement on the broader impact of the work, or other ethical considerations after the conclusion but before the references. The ethics statement will not count toward the page limit (8 pages for long, 4 pages for short papers).

\bibliography{custom}
\bibliographystyle{acl_natbib}

\clearpage
\appendix

\begin{figure}
    \centering
    \includegraphics[width=\linewidth]{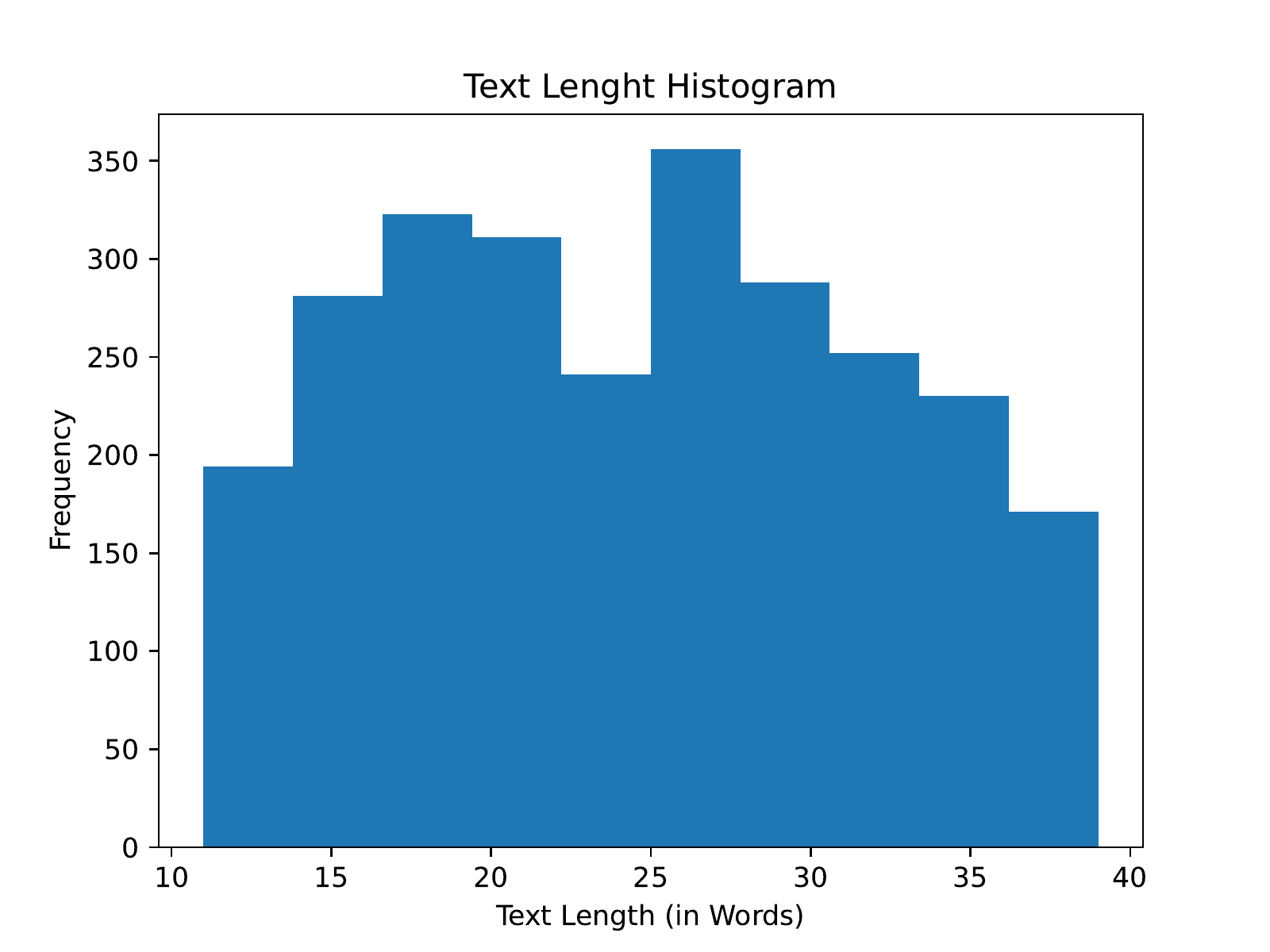}
    \caption{A Histogram of Text Lengths in our Dataset.}
    \label{tab:app_histogram_textlength}
\end{figure}

\section{Sample Selection}
\label{sec:sample_selection}
The basis for selecting samples are documents from four domains in the text produced by companies. We consider TCFD reports that are voluntarily self-disclosed by firms about their environmental impact, but not legally binding. Furthermore, we consider annual reports, comprehensive reports about activities conducted by a firm in a given year. We also consider corporate earnings calls (only the answer sections), which are conference calls between the management of a public company, analysts, investors, and the media to discuss the company’s financial results and other business-relevant topics during a given reporting period. Earnings conference call transcripts are obtained from Refinitiv Company Events Coverage (formerly Thomson Reuters StreetEvents). Lastly, we include the language data on environmental risks, targets, and performance from the CDP disclosure questionnaire responses from 2021. We denote the universe of these documents by $D_{\mathrm{large}}$. In Table \ref{tab:dataset_shares}, we show many sentences we have from each of these sources (first row), and the distribution of these sources in our final dataset (second row).

\begin{table}[h!]
    \centering
    \tiny
    \begin{tabular}{l | c c c c | c}
    Share & TCFD Reports & Annual Reports & CDP & Earning Calls & N \\ \hline
    All data &  0.07 & 0.20 & 0.00 & 0.73 & 16Mio \\
    Dataset & 0.21 & 0.41 & 0.01 & 0.37 & 2'647 \\ \hline
    \end{tabular}
    \caption{Data distribution over different sources (in \%), and sentence distribution in our dataset over different sources (in \%). The last column indicates the number of overall sentences.}
    \label{tab:dataset_shares}
\end{table}

In pilot studies, we decided to only keep sentences having more than 10 and less than 40 words. Shorter sentences rarely ever are environmental claims, but a combination of section titles, filler sentences, and table descriptions. Longer sentences usually are the result of a failure in preprocessing. 

A random selection of sentences from these documents would lead to a high number of sentences not related to the environment, thus, is impracticable. We also decided against using a keyword search to pre-filter $D_{\mathrm{large}}$ for two reasons. If we use a keyword set that is too narrow, we might have dataset artifacts. On the other hand, if we use a set that is too loose, we might again end up with too many non-climate-related sentences, which again is impracticable. 

As a remedy, we start with a handpicked selection of 250 environmental claims used in a recent marketing study about greenwashing in French investment funds by 2DII, an independent, non-profit think tank working to align financial markets and regulations with the Paris Agreement goals. We also consider 200 non-environmental claims as negative samples, randomly sampled from company websites. The authors translated them to English (if necessary) and loosely annotated these sentences to double-check their quality and to help come up with annotation guidelines. However, these 450 sentences do not appear in the final version of the dataset. Next, we train a preliminary RoBERTa\textsubscript{base} model on this dataset and use this trained model to compute the likelihood of each sentence from $D_{\mathrm{large}}$ being an environmental claim. Using this likelihood, we use the following strategy to select both samples with a high chance of being environmental claims, samples with a low chance of being environmental claims, and samples that are semantically similar but lead to very different results compared to our base transformer model:

\begin{enumerate}
    \item First 300 samples were sampled, which are adjacent to our starting selection of 250 environmental claims in SBERT embedding space \cite{reimers-gurevych-2019-sentence}, but for which the base transformer model assigned a small score of being an environmental claim.
    \item Then, 1500 samples with a score greater than 0.7 from our preliminary transformer model are selected.
    \item Next, 500 samples with a score between 0.2 and 0.5 from our preliminary transformer model are selected.
    \item Then, we selected 200 samples with a score lower than 0.2 from our preliminary transformer model.
    \item Finally, all encoded samples from SBERT are clustered into 2000 clusters using k-means. The largest clusters, from which no sample was selected in steps 1-4, are then represented by a random sample from the cluster. This way we increase the coverage of the whole domain by our selected samples. We selected 500 samples with that strategy.
\end{enumerate}

While we tried to maximize domain coverage using this sampling procedure, given the limited annotation budget, it is likely that we missed lots of utterances of environmental claims. Also, the sample is somewhat biased toward our preliminary model, which we used to sample environmental claims from. Moreover, we did not include all domains of text produced by listed companies. For example, company websites and advertisements are not included in our universe of documents.

\section{Annotation Guidelines}
\label{sec:annotation_guidelines}

Your task is to label sentences. The information we need is whether they are environmental claims (yes or no). 

A broad definition for such a claim is given by the European Commission: \textit{Environmental claims refer to the practice of suggesting or otherwise creating the impression [...] that a product or a service is environmentally friendly (i.e., it has a \textbf{positive impact} on the environment) or is \textbf{less damaging} to the environment than competing goods or services [...]}

In our case, claims relate to \textbf{products, services OR specific corporate environmental performance.}

\paragraph{General annotation procedure/principles :}
\begin{itemize}
    \item You will be presented with a sentence and have to decide whether the sentence contains an \textbf{explicit} environmental claim.
    \item Do not rely on implicit assumptions when you decide on the label. Base your decision on the information that is available within the sentence. 
    \item However, if a sentence contains an abbreviation, you could search online for the meaning of the abbreviation before assigning the label.
    \item In case a sentence is too technical/complicated and thus not easily understandable, it usually does not suggest to the average consumer that a product or a service is environmentally friendly and thus can be rejected.
    \item Likewise, if a sentence is not specific about having an environmental impact for a product or service, it can be rejected.
    \item Final goal: We will train a classifier on these annotations and apply it to massive amounts of financial text to explore which companies/sectors at which time make how many environmental claims. Does the number of environmental claims correlate with sectors/companies reducing their environmental footprint?
    \item The annotation task is not trivial in most cases. Borderline decisions are often the case. If you are uncertain about your decisions, copy-paste the sentence and add an explanatory note to the sentence. We will then cross-check it in case needed.
\end{itemize}

In Table \ref{tab:positive_annotation_examples} and \ref{tab:negative_annotation_examples}, we show examples that were discussed within the author team.

We presented each sentence in our sample to four annotators to determine a label. In the case of a clear majority of the annotators for a sentence (4:0, 3:1, 1:3, or 0:4), the sentence is annotated as such. In case of no majority (2:2), the sentence is discarded and excluded from our final dataset. The rationale behind this is that a sentence annotated as \emph{positive} accuses the writer to claim something. This accusation should be agreed on by the majority of readers (in dubio pro reo - in doubt, rule for the accused).
 
\storestyleof{description}
\begin{table*}[t]
\begin{listliketab}
{ \footnotesize
\begin{tabularx}{\linewidth}{@{}>{}l@{\hspace{.5em}}X@{} @{\hspace{.5em}}X@{}}
   Label & Sentence & Explanation \\ \hline  
    yes (unanimously) & Farmers who operate under this scheme are required to dedicate 10\% of their land to wildlife preservation. & Environmental scheme with details on implementation \\
    yes (borderline) & We prove our commitment to a sustainable world every day—by being a force for change where we work and live and holding ourselves and our suppliers to high standards in three vital aspects of doing business: people, product, and planet. & Very generic sustainability or responsibility wording without clear reference to environmental aspects. Yet the term “sustainability” and “responsibility” includes environmental aspects. \\ 
        yes (borderline) & Our places, which are designed to meet high sustainability standards, become part of local communities, provide opportunities for skills development and employment and promote wellbeing. & No would be: “Our places, which are designed to become part of local communities, provide opportunities for skills development and employment and promote wellbeing.” \\
  yes (borderline) & Fast Retailing has adopted ''Unlocking the Power of Clothing” for its Sustainability Statement, and through the apparel business seeks to contribute to the sustainable development of society. & Very generic sustainability or responsibility wording without clear reference to environmental aspects. Yet the term “sustainability” and “responsibility” includes environmental aspects. \\
    yes (borderline) & Hermès, which is currently managed by the sixth generation of family shareholders, is aware of its social responsibility and strives to give back to the world a part of what it gives to the Company. & Very generic sustainability or responsibility wording without clear reference to environmental aspects. Yet the term “sustainability” and “responsibility” includes environmental aspects. \\ 
    yes (borderline) & In 2016, UTC was placed on the CDP climate change and supplier A List, and in 2017 and 2018 received an A- and Leadership designation. & Environmental initiatives  and leadership. \\
    yes (borderline)& Change internal behavior; Drive low-carbon investment; Identify and seize low-carbon opportunities; Stakeholder expectations. & Intangible but environmentally friendly/ier processes. \\
    yes (borderline) & We are looking into the Insurance Underwriting element, and have taken part in the CRO Forum's Sustainability Carbon Footprinting paper of Underwriting. & Intangible but environmentally friendly/ier processes. \\
    yes (borderline) & In a further demonstration of the importance we place on helping customers to live sustainably, we became signatories of the Task Force on Climate related Financial Disclosures, to provide consistent information to our stakeholders. & Intangible but environmentally friendly/ier processes. \\ 
    yes (borderline) & As for assets, DBJ Green Building certification for 18 properties, BELS certification for 33 properties, and CASBEE certification for one property have been received. & Official environmental Labels \\
    yes (borderline) & Our clean, safe and high-tech products and solutions enable everything from food production to space travel, improving the everyday life of people everywhere. & Environmentally friendly/ier products and solutions \\
    yes (borderline) & FreshPoint, our specialty produce company, addresses customers' needs for fresh, unique, organic, and local produce items. & Environmentally friendly/ier products and solutions \\
    yes (borderline) & WilLDAR consists of detecting methane leaks with an optical gas imaging camera and repairing those leaks within 30 days. & Environmentally friendly/ier products and solutions \\
    yes (borderline) & These products include climate metrics, Climate Value-at-Risk (VAR), carbon portfolio reporting, low carbon, and climate change indexes as well as tools to identify clean-tech and environmentally oriented companies. & Environmentally friendly/ier products and solutions \\ \hline
\end{tabularx}
}
    \caption{Environmental Claims with Rationale in Annotation Guidelines}
    \label{tab:positive_annotation_examples}
\end{listliketab}
\end{table*}

\begin{table*}[t]
\begin{listliketab}
{ \footnotesize
\begin{tabularx}{\linewidth}{@{}>{}l@{\hspace{.5em}}X@{} @{\hspace{.5em}}X@{}}
   Label & Sentence & Explanation \\ \hline  
no (borderline) & We do this for 15 sustainable and impact strategies (equities, bonds and green bonds). & No positive impact or no link to better environmental performance \\
no (borderline) & We use the EcoAct ClimFIT (Climate Financial Institutions Tool) tool to measure the carbon emissions associated with the household and personal products sector. & No positive impact or no link to better environmental performance \\
no (borderline) & AUSEA is a miniaturized sensor, fitted onto a commercial drone, that can detect methane and carbon dioxide. & Product with potentially positive environmental impact, but impact is not stated hence no claim \\
no (borderline) & This will further accelerate Croda's positive impact by creating and delivering solutions to tackle some of the biggest challenges the world is facing. & Unclear whether this relates to environmental positive impacts, only implicit assumptions would make it an environmental claim. \\
no (unanimously) & Hence, the Scope 2 emission is included in the Scope 1 emission which has been reported in accordance with the ISO 14064-1 requirements as verified by qualified independent assessor. & Technical details, descriptions, and explanations \\
no (unanimously) & Emissions associated with processing activities are associated with the supply of these ingredients and are included in our Scope 3 supply chain emissions. & Technical details, descriptions, and explanations \\
no (unanimously) & Emissions are modelled based on sector averages including linear regression and country carbon emissions intensities for GDP. & Technical details, descriptions, and explanations \\
no (unanimously) & Wood products facilities also operate lumber drying kilns and other processes that can either use the steam from the boilers or, if direct fired, will commonly use natural gas. & Technical details, descriptions, and explanations \\
no (unanimously) & We use the EcoAct ClimFIT (Climate Financial Institutions Tool) tool to measure the carbon emissions associated with utilities. & Technical details, descriptions, and explanations \\
no (unanimously) & In the past we have conducted analysis of our portfolio impact on the climate, using scope 3 as a metric. & Technical details, descriptions, and explanations \\
no (unanimously) & For that, Danone needs organic fresh milk. & Sentence context would be required to understand whether it is a claim \\
no (unanimously) & UPM Biofuels is developing a new feedstock concept by growing Brassica Carinata as a sequential crop in South America. & Sentence context would be required to understand whether it is a claim \\
no (unanimously) & Our key sources of emissions are the running of our operations (electricity, business travel, etc), purchased goods and services (consultants, maintenance work, IT services, etc), and land leased to sheep and beef farming (to keep the grass low under our wind farms). & environmental risk exposure description but no commitment / claim to act on reducing the risk or improving impact \\
no (unanimously) & Extreme weather events and the impacts of transitioning to a low-carbon economy have the potential to disrupt business activities, damage property, and otherwise affect the value of assets, and affect our customers' ability to repay loans. & environmental risk exposure description but no commitment / claim to act on reducing the risk or improving impact \\
no (unanimously) & At the date of this report, the Group owns 34 mills (29 of which produce containerboard), 245 converting plants (most of which convert containerboard into corrugated boxes), 40 recovered fibre facilities and two wood procurement operations (which together provide raw material for our mills) and 34 other production facilities carrying on other related activities. & environmental risk exposure description but no commitment / claim to act on reducing the risk or improving impact \\
%no (unanimously) & In addition to fresh and frozen salmon, Mowi offers a wide range of value-added products ranging from whole gutted fish, through products such as fillets, steaks and portions, to smoked salmon and ready-to-eat dishes. & environmental risk exposure description but no commitment / claim to act on reducing the risk or improving impact \\
%no (unanimously) & However there remains uncertainty in the industry around the right long-term solutions for homes due to changing energy policies and uncertainty around carbon emissions from grid electricity and gas in the future. & environmental risk exposure description but no commitment / claim to act on reducing the risk or improving impact \\
%no (unanimously) & We are committed to conducting business according to the highest ethical standards and in compliance with the law. & Just comply with law and regulations and no additional action \\
%no (unanimously) & Norfolk Southern's operations are subject to federal and state environmental laws and regulations concerning, among other things, emissions to the air; discharges to waterways or ground water supplies; handling, storage, transportation, and disposal of water and other materials; and the clean-up of hazardous material or petroleum releases. & Just comply with law and regulations and no additional action \\ 
\hline
\end{tabularx}
}
    \caption{Negative Examples with Rationale in Annotation Guidelines}
    \label{tab:negative_annotation_examples}
\end{listliketab}
\end{table*}

\section{Data Sources}\label{app:data_sources}

\begin{figure}[h!]
\centering
\begin{subfigure}{\linewidth}
    \centering
    \includegraphics[width=\linewidth]{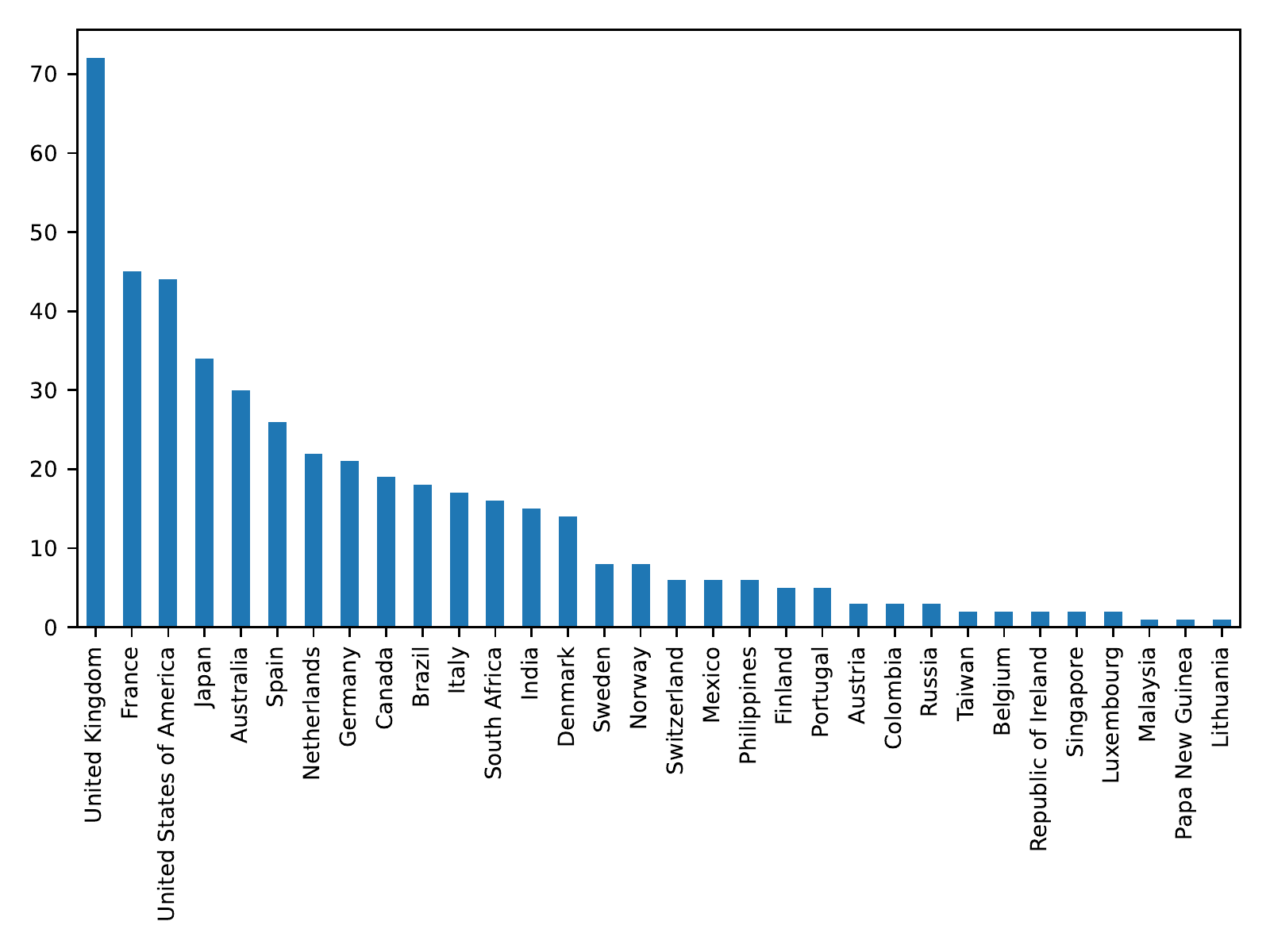}
    \caption{Geographical Distribution for annual and TCFD Reports in our Data.}
    \label{tab:countries_distribution}
\end{subfigure}

\begin{subfigure}{\linewidth}
    \centering
    \includegraphics[width=\linewidth]{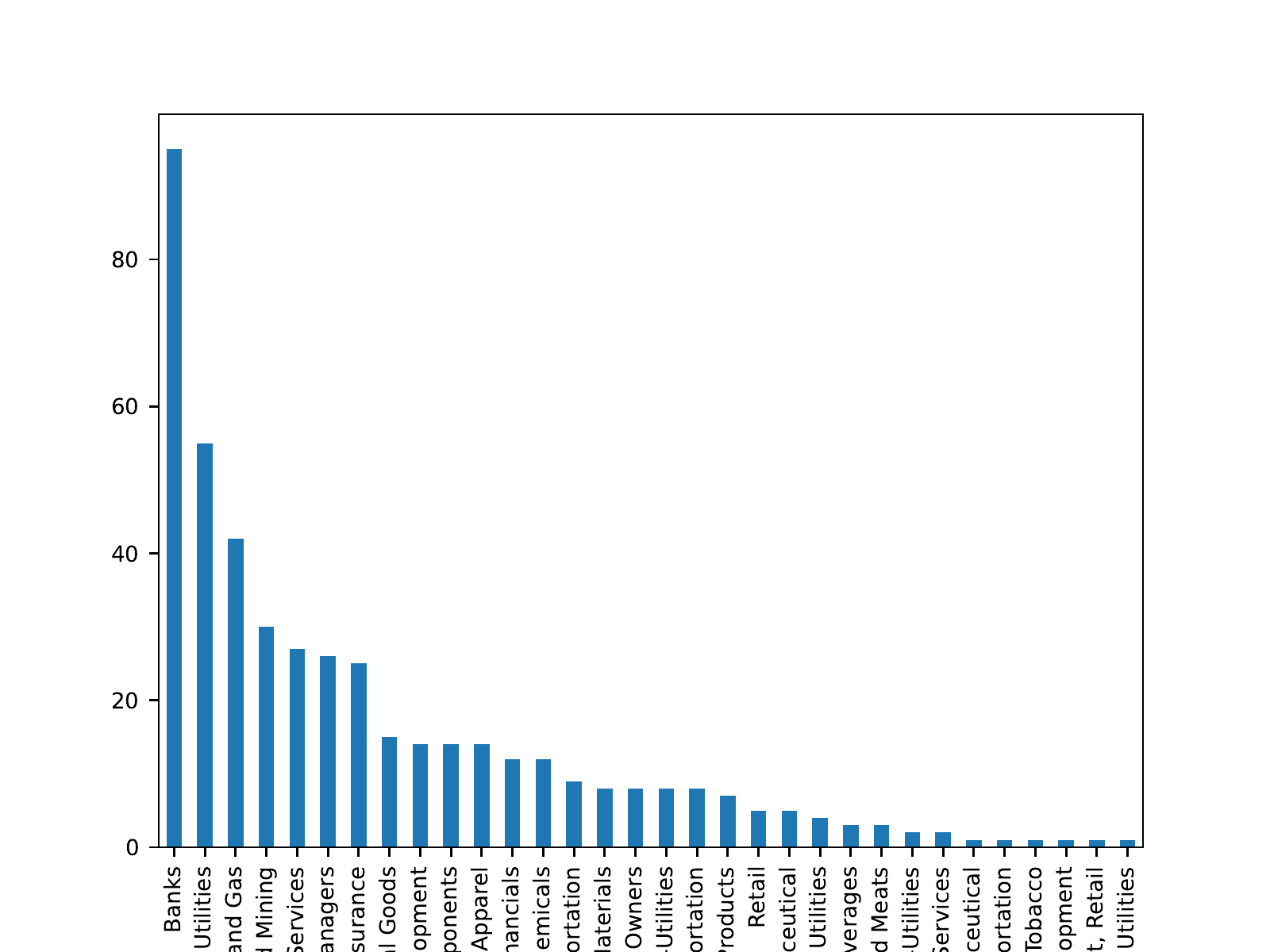}
    \caption{Distribution over Sectors for annual and TCFD  Reports in our Data.}
    \label{tab:industry_distribution}
\end{subfigure}

\begin{subfigure}{\linewidth}
    \centering
    \includegraphics[width=\linewidth]{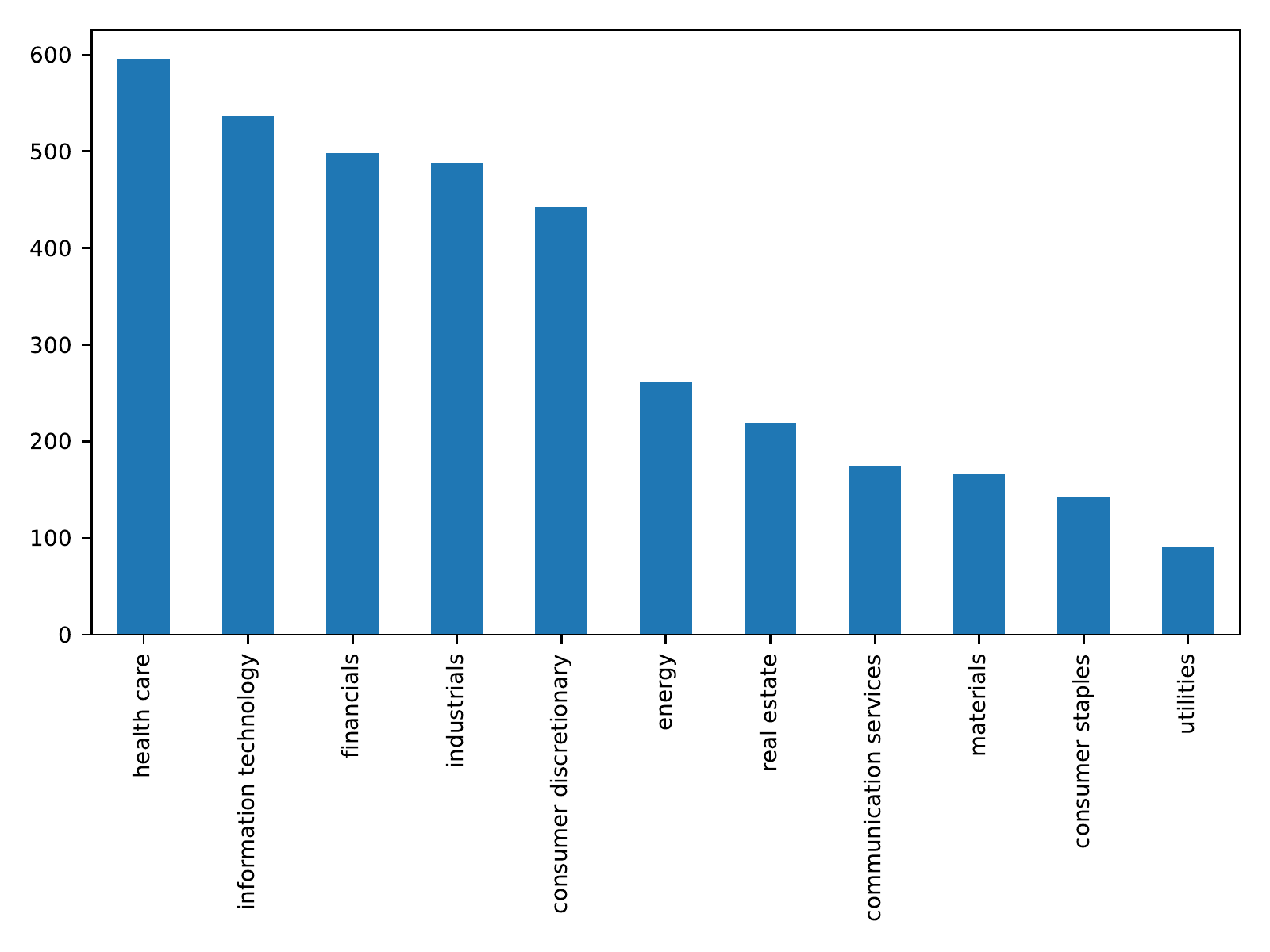}
    \caption{Distribution over Sectors for quarterly earning calls in our Data.}
    \label{tab:industry_sector}
\end{subfigure}
\end{figure}

We crawled TCFD and annual reports from the SEC (the U.S. Securities and Exchange Commission), specifically from \url{www.annualreports.com} \url{www.responsibilityreports.com}. Given that sustainability reports are mostly published by European and US firms, there is not an even global coverage in our sample, but a tendency for firms in developed countries. For the reports we collected, we show a distribution of Countries in Figure \ref{tab:countries_distribution} and Industries in Figure \ref{tab:industry_distribution}. For the earning calls data, we show a distribution over sectors in Figure \ref{tab:industry_sector}.

%We crawled the reports from companies from the following sectors: \textit{Banks, Insurance, Retail, Oil and Gas, IT and Communications Services, Electric and Gas Utilities, Automobiles and Components, Asset Managers, Metals and Mining, Paper, Packaging and Forest Products, Consumer Goods and Apparel, Construction Materials, Chemicals, Capital Goods, Diversified Financials, Construction Materials, Consumer Goods and Apparel, Real Estate Management and Development, Retail, Chemicals, Electric and Gas Utilities, Real Estate Management and Development, Multi-Utilities, Beverages, Pharmaceutical, Packaged Foods and Meats, Commercial and Professional Services, Packaged Foods and Meats, Pharmaceutical, Water Utilities, Maritime Transportation, Passenger Air Transportation, Rail Transportation, Tobacco, Diversified Financials, IT and Communications Services, Multi-Utilities, Real Estate Management and Development, Asset Owners, Consumer Goods and Apparel, Diversified Financials, Multi-Utilities.}

%Roughly half of the reports are coming from the following sectors: \textit{Banks, Electric and Gas Utilities, Oil and Gas, Metals and Mining, IT and Communications Services.}

\section{Dataset Size}\label{sec:dataset_size}

\begin{figure}
    \centering
    \includegraphics[width=\linewidth]{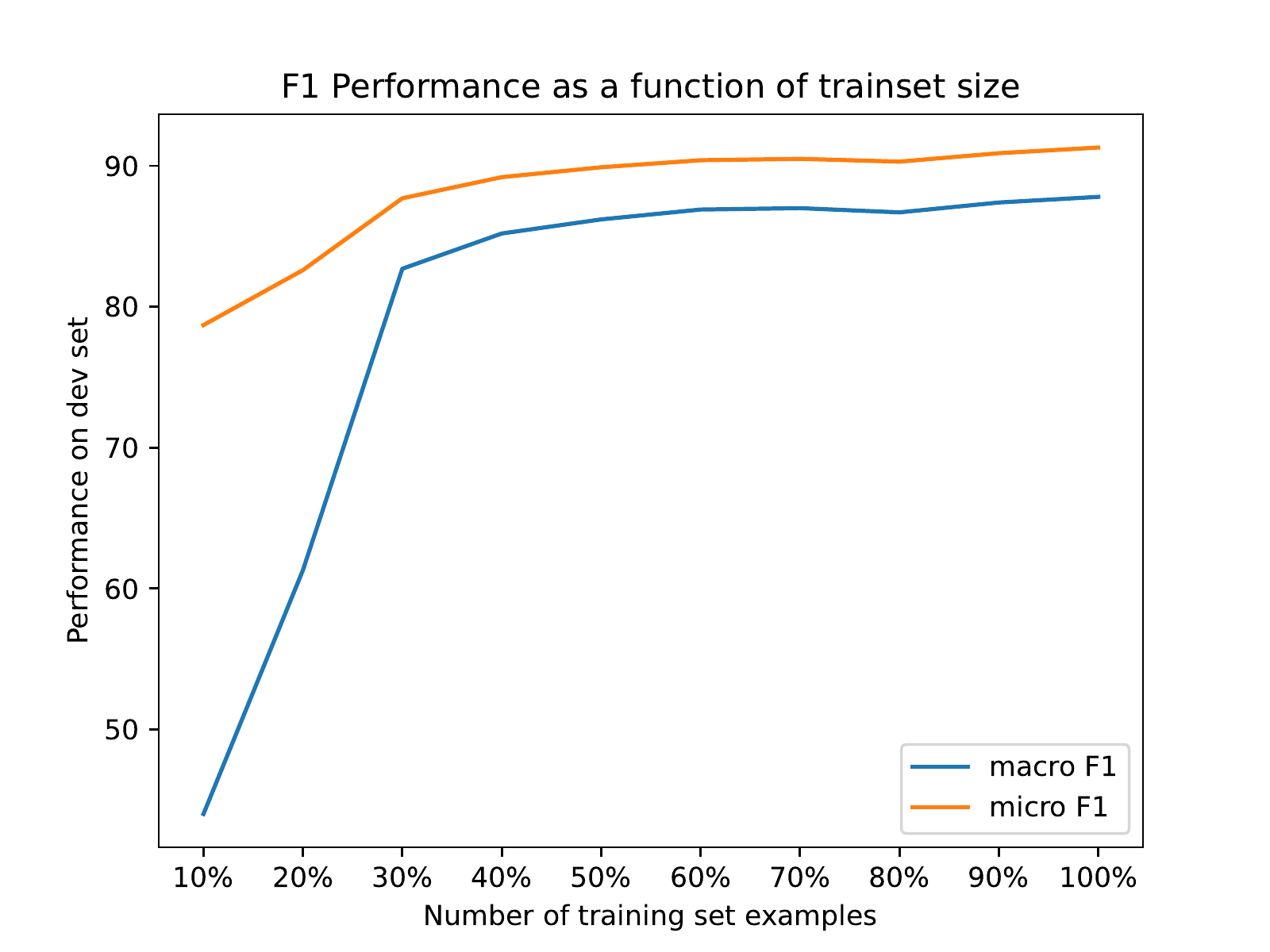}
    \caption{Performance of ClimateBERT on the development set as a function of training on different fractions of the training dataset.}
    \label{fig:dataset_size}
\end{figure}

Figure \ref{fig:dataset_size} shows that model performance as a function of dataset size converges quickly. We fine-tune a ClimateBERT model on different subsets of the training data, e.g. on 10\%, on 20\%, etc. In Figure \ref{fig:dataset_size}, we find diminishing marginal utility after having fine-tuned a model on more than 60\% of the dataset. Hence, we believe that our dataset is sufficient in size and we do not expect model performance to increase drastically anymore if we were to annotate more data points.

\begin{table*}[]
    \centering
    \footnotesize
    \begin{tabular}{p{7.5cm} p{7.5cm}}
    Environmental Claims & Negative Examples \\ \hline
    In support of Apple's commitment to reduce its carbon footprint by transitioning its entire supply chain to 100\% renewable energy, we've transitioned our facilities in China to be powered through a series of renewable power purchase agreements. & So there's an annual cycle that, to some degree, dictates the pace of these enrollment campaigns. \\
    We are looking at opportunities to expand our commitment to renewable diesel while continuing to optimize the efficiency of our fleet of traditional biodiesel plants. & And so when we get these biopsy data published, which we're aggressively working on, we think we will have sufficient information to begin to approach payers, including Medicare. \\
    We plan to continue our low risk growth strategy by building our core business with rate base infrastructure, while maintaining the commitment to renewable energy initiatives and to reducing emissions. & And I guess first of all, I would say the thesis which we have at FERC here for precedent is no different than what takes place right now for the LDC companies, where the LDC companies pay for pipeline infrastructure that's developed by a pipeline operator. \\
    We just completed \$1 billion of capital projects to expand, upgrade and modernize and improve the environmental footprint of an important industry in Russia. & But as Jon points out, the thing that they really seem to be focused on is we claim a five-year life, and they want to make sure that that's a reasonable claim on our batteries for AED Plus. \\
    And we also announced that BHGE is committed to reduce its carbon footprint by 50\% by 2030, and also net 0 by 2050. & They're critical to reimbursement, meaning you just simply can't get revenue unless you've done things like enroll it, and you have to have accurate data to get providers enrolled. \\ \hline
    \end{tabular}
    \caption{Environmental Claims and Negative Examples Predicted in Quarterly Earning Calls Answer Sections.}
    \label{fig:examples_earning_calls}
\end{table*}

\section{Environmental Impact}

In this section, following \cite{towardsclimateawareness} we describe the environmental impact of  our dataset construction and experiments. All experiments were conducted on a carbon-neutral computing cluster in Switzerland, using a single Nvidia GeForce GTX 1080 Ti GPU with a TDP of 250 W., see \url{https://www.hpc-ch.org/lake-water-to-cool-supercomputers-at-cscs}. While the computing cluster we performed the experiments on is superficially carbon-neutral, there are still emissions for the production and shipping of the hardware used. Also, the energy used for our experiments could replace power produced by fossil fuel somewhere else. Therefore, we calculate emissions based on the country's energy mix.

Running the main experiments took less than 1 hour combined. Detecting environmental claims in the quarterly earning calls took an additional 3 hours. For preliminary experiments, we trained a battery of transformer models on loosely annotated data (we used scores assigned by our "best" model to sample the sentences in the dataset). This took roughly 48 hours. Also, we embedded all sentences with SBERT for two additional hours. In total, we spent about 60 hours of computation time.

\begin{table*}[ht]
\scriptsize
\begin{tabular}{p{6cm}p{9cm}}
\hline
\multicolumn{2}{c}{\textbf{Minimum card}} \\
\hline
\textbf{Information} & \textbf{Unit} \\
\hline
1. Is the resulting model publicly available? & \textit{upon publication} \\\\
2. How much time does the training of the final model take? & $<5$ min \\\\
3. How much time did all experiments take (incl. hyperparameter search)? & 60 hours \\\\
4. What was the energy consumption (GPU/CPU)? & 0.3 kW \\\\
5. At which geo-location were the computations performed? &
\textit{Switzerland} \\
\hline
\multicolumn{2}{c}{\textbf{Extended card}} \\
\hline
6. What was the energy mix at the geolocation? & 89 gCO2eq/kWh \\\\
7. How much CO2eq was emitted to train the final model? & 2.2 g \\\\
8. How much CO2eq was emitted for all experiments? & 1.6 kg \\\\
9. What is the average CO2eq emission for the inference of one sample? & 0.0067 mg \\\\
10. Which positive environmental impact can be expected from this work? & This work can help detect and evaluate environmental claims and thus have a positive impact on the environment in the future. \\\\
11. Comments & -  \\

\end{tabular}
\caption{Climate performance model card following \cite{towardsclimateawareness}}
\label{tbl:model_card_example}
\end{table*} 

\section{Funding}
{This paper has received funding from the Swiss National Science Foundation (SNSF) under the project (Grant Agreement No. 207800).}

\end{document}